\documentclass[letterpaper, 10 pt, conference]{ieeeconf}

\usepackage[normalem]{ulem}
\useunder{\uline}{\ul}{}
\usepackage{graphicx}
\usepackage{amsmath}
\usepackage{amssymb}
\usepackage{booktabs}
\usepackage[table,xcdraw, dvipsnames]{xcolor}

\usepackage{tabularx}
\usepackage{multirow}
\usepackage{siunitx}
\usepackage{subcaption}
\usepackage[font=small]{caption}

\usepackage{balance}
\usepackage{url}
\usepackage{mathtools}
\usepackage{siunitx}
\usepackage[acronym]{glossaries}

\usepackage{amsmath}

\usepackage{bm}

\usepackage{cite}

\usepackage{color}
\definecolor{mygreen}{RGB}{26, 148, 49}

\newcommand{\reffig}[1]{Fig.~\ref{#1}}

\usepackage{float}
\usepackage[multiple]{footmisc}

\IEEEoverridecommandlockouts   

\begin{document}
\title{\LARGE \bf CHORAL: Traversal-Aware Planning for Safe and Efficient Heterogeneous Multi-Robot Routing}
% TraHTra: Traversal-Aware Routing for Heterogeneous Teams of Robots for Safe and Efficient Inspection

\author{
David Morilla-Cabello, and Eduardo Montijano%
\thanks{This work was partially funded by Spanish grant FPU20-06563, project T45\_23R, by grants AIA2025-163563-C31, PID2024-159284NB-I00 funded by MCIN/AEI/10.13039/501100011033 and ERDF, and the Office of Naval Research Global grant N62909-24-1-2081. D. Morilla-Cabello and E. Montijano are with the Instituto de Investigaci\'on en Ingenier\'ia de Arag\'on, Universidad de Zaragoza, Spain.
Corresponding: \texttt{\small davidmc@unizar.es}}
}

% make the title area
\maketitle

\begin{abstract}

Monitoring large, unknown, and complex environments with autonomous robots poses significant navigation challenges, where deploying teams of heterogeneous robots with complementary capabilities can substantially improve both mission performance and feasibility. However, effectively modeling how different robotic platforms interact with the environment requires rich, semantic scene understanding.
Despite this, existing approaches often assume homogeneous robot teams or focus on discrete task compatibility rather than continuous routing. Consequently, scene understanding is not fully integrated into routing decisions, limiting their ability to adapt to the environment and to leverage each robot’s strengths.
In this paper, we propose an integrated semantic-aware framework for coordinating heterogeneous robots. Starting from a reconnaissance flight, we build a metric-semantic map using open-vocabulary vision models and use it to identify regions requiring closer inspection and capability-aware paths for each platform to reach them. These are then incorporated into a heterogeneous vehicle routing formulation that jointly assigns inspection tasks and computes robot trajectories.
Experiments in simulation and in a real inspection mission with three robotic platforms demonstrate the effectiveness of our approach in planning safer and more efficient routes by explicitly accounting for each platform’s navigation capabilities. We release our framework, CHORAL, as open source to support reproducibility and deployment of diverse robot teams.

\end{abstract}

%%%%%%%%% BODY TEXT
\section{Introduction} 
\label{sec:intro}

\begin{figure}[!t]
    \centering
    \includegraphics[width=\linewidth]{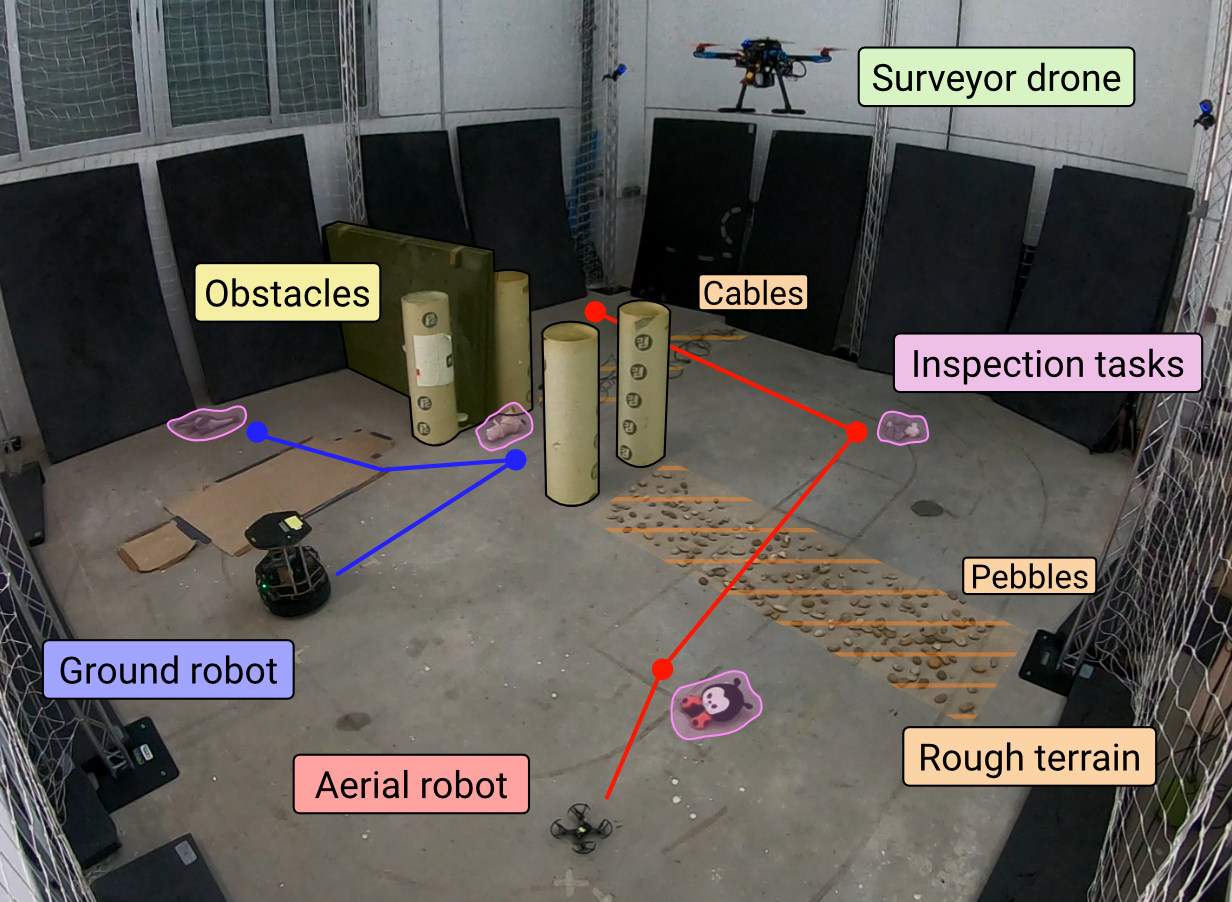}
    \caption{Our framework for heterogeneous multi-robot inspection in a real-world deployment. A metric-semantic map of the environment is built from a \textcolor{YellowGreen}{reconnaissance flight}, enabling the automatic identification of \textcolor{Lavender}{inspection targets} and the computation of platform-specific, traversal-aware routes. \textcolor{blue}{Ground robots} avoid regions with \textcolor{orange}{poor traversability} with pebbles or cables, while \textcolor{red}{aerial robots} plan safer trajectories that maintain clearance from \textcolor{Goldenrod}{obstacles}.}
    \label{fig:teaser}
\end{figure}

Modern mobile robotic platforms, including wheeled, legged, and aerial systems, can now operate with remarkable autonomy, each offering distinct mobility and endurance characteristics~\cite{morais2025fleetsurvey, queralta2020applicationshetmarine}. 
Leveraging these complementary strengths in heterogeneous teams has the potential to expand the scale, safety, and efficiency of missions such as search and rescue, environmental monitoring, and large-area industrial inspection~\cite{tranzatto2024cerberus, malladi2025digiforest}. 
In these scenarios, robots must not only cover vast and often unknown environments, but also identify which regions merit fine-grained inspection and determine which robot is best suited to visit them  (\reffig{fig:teaser}).

Despite this need, existing high-level coordination strategies address only parts of the problem. Classical routing approaches, such as the Vehicle Routing Problem (VRP), environment decomposition, or iterative allocation~\cite{david2022sweep,zhou2023racer, hardouin2023centdec} typically assume homogeneous teams and optimize only total distance or completion time.
Conversely, task-assignment methods that incorporate compatibility or precedence constraints~\cite{notomista2022maintaskassigment, fu2023robustscheduling} treat routing simplistically and do not reason about how terrain, obstacles, or platform-specific mobility affect accessibility. Moreover, existing approaches do not integrate semantic scene understanding, limiting their applicability to complex and realistic inspection missions involving heterogeneous platforms.

Recent advances in scene understanding have enabled robots to build rich metric–semantic representations of their environment. Open-vocabulary vision models can reason about semantic concepts involving terrain, obstacles, or relevant objects~\cite{frey23fast, yoon2024travcnn,fortin2025collaborativenav}.
However, despite the growing availability of such flexible perceptual tools, their direct integration into multi-robot planning frameworks remains largely unexplored~\cite{notomista2022maintaskassigment,miller2022strongertogether,cai2023energyaware}.

In this work, we introduce a semantic-aware coordination framework, CHORAL, which unifies environment understanding and heterogeneous routing for multi-robot inspection. Our system begins with a reconnaissance aerial survey from which we construct an open-vocabulary metric–semantic map using state-of-the-art vision models. This representation enables the automatic identification of inspection regions while providing terrain context for planning that reflects each robot’s navigation capabilities.
We then formulate a Heterogeneous Vehicle Routing Problem that explicitly incorporates traversal information into robot-specific path costs, allowing the planner to leverage platform heterogeneity to compute safe and efficient task assignments and routes.

\begin{figure*}[!ht]
    \centering
    \vspace{3mm}
    \includegraphics[width=0.95\textwidth]{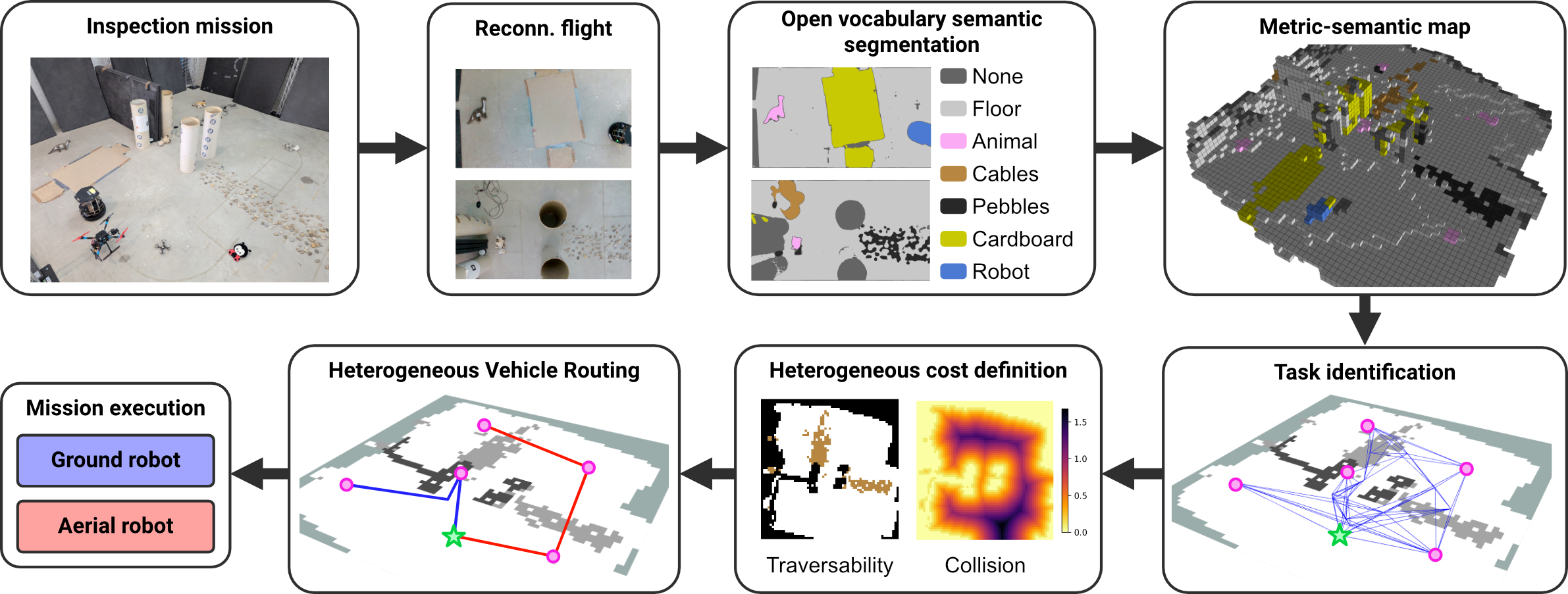}
    \caption{Overview of our framework for heterogeneous multi-robot routing. (a) An inspection mission is defined in an unknown environment based on user-specified task classes. (b) A surveying aerial platform constructs a metric-semantic map using open-vocabulary vision models. (c) The map is processed to identify inspection targets, compute feasible connections between tasks, and extract platform-specific traversal costs. (d) A heterogeneous vehicle routing problem is formulated and solved to generate safe and efficient routes tailored to the capabilities of each robot. (e) The resulting plans are executed using standardized navigation stacks.}
    \label{fig:diagram}
    \vspace{-4mm}
\end{figure*}

We evaluate our framework in simulated environments of varying size and complexity, and we further demonstrate real-world deployment using three robotic platforms in a ROS 2-based inspection mission. 
Our experiments demonstrate that our proposed framework obtains safer and more efficient routes for each robot. Additionally, we validate the applicability of our system to a real scenario, being able to model the environment and plan accounting for varied platforms.
The full system, including open-vocabulary semantic mapping and heterogeneous route planning, is containerized and released as open source\footnote{Software will be released upon acceptance}, supporting reproducibility and accelerating research on large-scale heterogeneous multi-robot coordination.

\section{Related Work}
\label{sec:related}

Deploying multi-robot teams for inspection, exploration, and mapping improves operational efficiency in large environments~\cite{morais2025fleetsurvey, queralta2020applicationshetmarine}, but introduces significant challenges in routing and coordination. Scene-reconstruction and inspection pipelines often adopt hierarchical strategies that partition the environment and assign viewpoints through greedy allocation, followed by refinement via Traveling Salesman Problem (TSP)~\cite{hardouin2023centdec} or Vehicle Routing Problem (VRP) variants. These include min–max and capacitated VRPs~\cite{david2022sweep, zhou2023racer}, consistent multiple depot multiple TSP~\cite{zhang2024soar}, and efficient graph-based approaches for cluttered spaces~\cite{xu2025efficient}. While effective for structured inspection tasks, these methods assume homogeneous teams and rely primarily on distance-based costs, limiting their ability to exploit differences in robot dynamics.

Heterogeneous coordination is increasingly studied due to the complementary strengths of aerial and ground platforms~\cite{queralta2020applicationshetmarine, miller2022strongertogether, li2024cooperativeexplo}, as demonstrated in the DARPA Subterranean Challenge~\cite{tranzatto2024cerberus}. However, routing in such contexts remains largely restricted to simplified task assignment or environment partitioning rather than fine-grained, cost-aware planning. In parallel, the Heterogeneous VRP introduces vehicle-dependent costs~\cite{golden1984hvrp} and has been extensively explored in logistics through heuristics, tabu search, and genetic algorithms~\cite{desrochers1991solutionhvrp, lai2016tabusearchhvrp, liu2009gahvrp}. These methods primarily target mixed-fleet transportation systems~\cite{goeke2015electrichvrp, vazPenna2016electrichvrp2}, with limited adoption in robotics.

In robotics, heterogeneous coordination is predominantly framed as a task assignment problem, emphasizing task–platform compatibility and resilience~\cite{notomista2022maintaskassigment, mao2025constraints}. While heuristic methods~\cite{calvo2022firstcomefirstserve}, Mixed-Integer Programs~\cite{gosrich2025onlineprecedence}, and learning-based approaches~\cite{dai2025taskassRL} provide effective allocation of individual tasks, they typically simplify or defer the routing component that is critical in inspection missions. Moreover, these formulations often assume known task compatibilities or rely on predefined energy maps, and are primarily validated. %in simulation or small-scale experiments.

Beyond task assignment, several works attempt to incorporate routing considerations more explicitly. Some combine allocation with distance-based costs~\cite{park2023tolerantassignment} or integrate platform-specific energy consumption into route planning~\cite{fu2023robustscheduling, caballero2024fixedmultihvrp, cai2023energyaware}. Others account for platform size to ensure access in constrained environments~\cite{cihlarova2024hetindoorexpl} or formulate cooperative UAV–UGV routing through compatibility constraints rather than full cost modeling~\cite{mathew2015ugvuavtogether}. These approaches, however, typically depend on tailored optimization problems, limited-scale validation, or restrictive assumptions, standing in contrast to the robust, industry-grade routing frameworks widely used in inspection~\cite{helsgaun2017lkh, ortools}.

Environmental awareness is essential for safe and feasible routing, as terrain constrains ground platforms and obstacles pose risks to aerial robots. Novel deep learning models for extracting language aligned features from images have shown impressive results in building open vocabulary models to obtain rich environmental awareness~\cite{shi2025trident,alama2025rayfronts}. Prior work models risk using metric maps~\cite{hakobyan2019riskcollision}, probabilistic terrain analysis based on geometry and using CVaR for modeling costs~\cite{dixit2024step}, or perception-based traversability estimation using semantics and self-supervision~\cite{frey23fast, yoon2024travcnn}, with collaborative estimation explored across UAVs and UGVs~\cite{fortin2025collaborativenav}. However, these techniques are applied to individual robot navigation. Furthermore, recent advances in deep learning have enabled the extraction of language-aligned features from images, providing rich semantic understanding of the environment~\cite{shi2025trident, alama2025rayfronts}. Our work integrates open-vocabulary environmental awareness directly into the routing process, enabling heterogeneous multi-robot systems to plan inspection missions that account for platform dynamics and environmental constraints, improving mission efficiency and feasibility.

\section{Semantic-aware Heterogeneous Coordination Framework}

Our framework for heterogeneous multi-robot coordination is illustrated in Fig.~\ref{fig:diagram}. An initial reconnaissance flight is used to construct an open-vocabulary metric-semantic map, which flexibly integrates user-defined inspection tasks, environment characteristics relevant to platform traversability, and obstacle information. This map enables both the identification of inspection targets and the estimation of platform-specific traversal costs. These elements are then incorporated into a heterogeneous vehicle routing problem formulation that jointly assigns tasks and computes routes, yielding safe and efficient plans that explicitly leverage the complementary capabilities of each robot in the team.

\subsection{Metric-semantic map construction}

Our approach begins with a reconnaissance flight over the unknown area to construct an open-vocabulary metric–semantic map.
This is performed by a localized aerial platform that surveys the environment along a fixed or exploratory trajectory, ensuring full coverage.
We leverage and adapt several off-the-shelf tools, described in detail in Section~\ref{sec:implementation}, to build a 3D voxel-based representation of the environment.

\begin{figure}[!t]
    \centering
    \vspace{2mm}
    \includegraphics[width=0.85\linewidth]{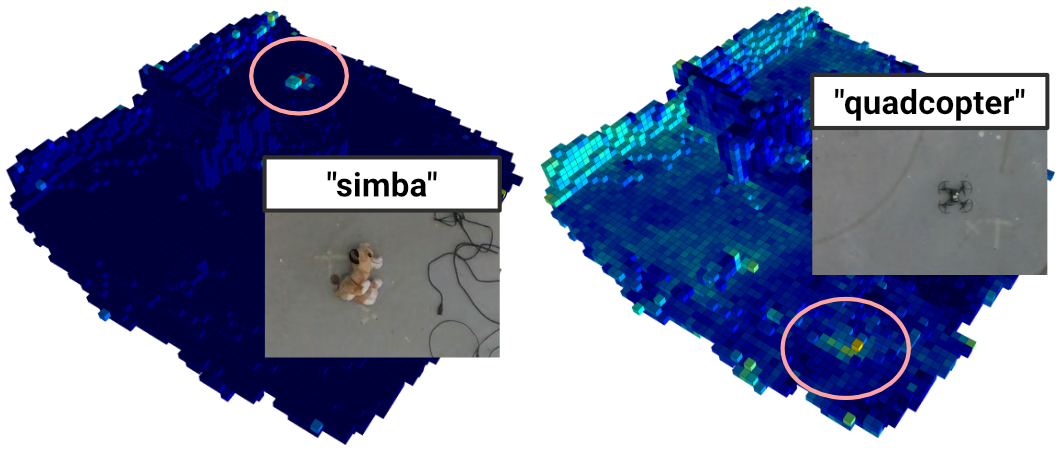}
    \caption{Open vocabulary map built with our mapping module, showing the flexibility in specifying semantically rich classes.}
    \label{fig:map}
\end{figure}

Each voxel contains geometry, represented as the probability of occupancy, and semantic information, encoded as a vector of features extracted using an open-vocabulary vision model. Our method supports two types of semantic representations: \emph{(i)} a categorical probability vector $\Gamma=\{\gamma_1, \dots, \gamma_n\}$ over an arbitrary set of semantic classes predefined by user prompts (e.g., grass, path, road, mud), see~\reffig{fig:diagram}; and \emph{(ii)} text-aligned raw semantic embeddings, which provide richer information online at the cost of increased memory usage, see~\reffig{fig:map}. We assume that the user prompts are divided into two categories, tasks $\mathcal{T}$ and environment characteristics $\mathcal{E}$, which will be used later on to identify places to inspect and the cost to reach them.

\begin{figure}[!t]
    \centering
    \includegraphics[width=0.95\linewidth]{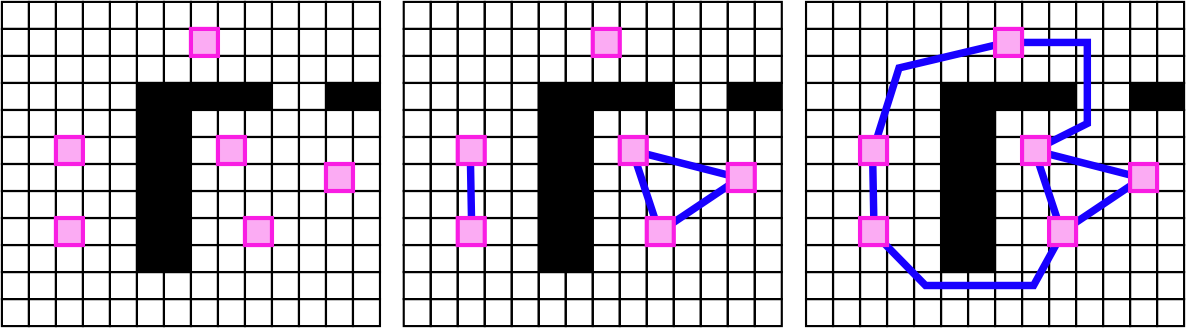}
    \caption{PRM construction. From left to right: the tasks are shown in blue and obstacles in black, the covisible tasks are connected, disjoint sets are detected and merged using RRT*.}
    \label{fig:planner}
\end{figure}

For efficient planning, we process the 3D map into a 2D grid representation. The semantic information of each 3D voxel is reduced to a categorical label based on the set of user-defined text queries. Raw semantic embeddings are assigned to the class whose text embedding has the largest cosine similarity. Categorical probabilities are assigned the label with the highest probability, $\arg\max\Gamma.$ 

The voxel labels are then collapsed vertically onto the 2D grid using a precedence rule: voxels above a height threshold are treated as obstacles; otherwise, task-related classes take priority, followed by environment characteristics that impede traversability, and finally free space. Each resulting 2D cell, $m\in\mathcal{M},$ stores a semantic label, $\gamma(m),$ and the distance to the nearest obstacle $d(m),$ computed from an Euclidean Distance Field (EDF) of the grid.

\subsection{Task identification}

Task-related semantic classes are used to identify regions in the 2D map that require fine-grained inspection. Individual inspection tasks are extracted by clustering grid cells classified as tasks using DBSCAN~\cite{kriegel96dbscan}, followed by outlier removal based on a minimum cluster size. The centroid represents a single task point. An inspection point is selected as the point along the robot’s inspection path that is closest to the task point while maintaining a desired separation distance for safety.

Let $\mathcal{V}$ denote the resulting set of inspection tasks. For each pair of inspection tasks, $i$ and $j$, we generate a path, $\pi_{ij}$  with an efficient, fully connected Probabilistic Roadmap (PRM), as illustrated in~\reffig{fig:planner}. 
First, all mutually visible tasks are connected with straight-line edges when no obstacles obstruct the line of sight. 
Next, disconnected components are identified using a flood-fill algorithm. 
To connect these components, we employ RRT* to find feasible paths between the closest tasks in disjoint sets. 
This procedure ensures that every task is reachable from any other through free space. 
Finally, A* search is applied to compute the shortest paths between all task pairs. 
Note that paths might potentially be composed of multiple PRM segments.
Our experiments show that this PRM construction is efficient even in large, cluttered environments. 
Moreover, maintaining a PRM structure allows for fast updates, such as node repositioning or edge recomputation, in response to changes in the environment.

\subsection{Heterogeneous costs definition}

We consider a team of $k\in K$ different robots for the fine-grained inspection of the tasks. By combining information from the metric–semantic map with each platform’s capabilities, we define traversal costs that accurately reflect the suitability of each robot for navigating between inspection targets.
The cost for robot $k$ to go from task $i$ to task $j$ is defined as
\begin{equation}
c_{ij}^{k} = c_{\text{time}}(\pi_{ij},k) + \alpha c_{\text{safe}}(\pi_{ij},k),
\end{equation}
where $\alpha$ balances the importance of safety relative to traversal time.

\subsubsection{Time cost}
The time cost captures both the path geometry and the dynamic agility of each platform. 
While it could also account for maneuvering requirements such as speed reductions along sharp turns, we approximate it using the nominal velocity of each robot, $v^k$. Denoting the path length of $\pi_{ij}$ by $d_{ij}$, the traversal cost is defined as
\begin{equation}
c_{\text{time}}(\pi_{ij},k) = \frac{d_{ij}}{v^k}.
\end{equation}

\subsubsection{Safety costs}
We model the safety costs via a statistical survival formulation along the path, assuming accidents follow a Poisson process (i.e., independent and infrequent) with respect to distance.
We use $\mathbf{x}^k \in \mathcal{X}$ to denote the state of robot $k$, including its position and characteristics. We also define $M(\mathbf{x}^k) : \mathcal{X} \to \mathcal{M}$ as the transformation from robot position to grid cell, denoted as $m^k\in\mathcal{M}$ for brevity. The probability that robot $k$, successfully traverses $\pi_{ij}$ is
\begin{equation}
c_{\text{safe}}(\pi_{ij}, k) = 1 - \exp\left(- \sum_{\mathbf{x}^k\in\pi_{ij}} \lambda(\mathbf{x}^k, m^k) \right).
\end{equation}
where $\lambda(\mathbf{x}^k, m^k)$ is the platform-specific accident rate per unit distance in a certain map cell. The accident rate depends on both environmental factors (e.g., terrain type, proximity to obstacles) and the robot’s specific characteristics.

To account for platform-specific risks associated with their motion in the environment, we model the expected accident rate of robot $k$ visiting $m$, as a function of environmental features derived from the metric-semantic map. 
In particular, our formulation combines traversability and collision risks,
\begin{equation}
\lambda(\mathbf{x}^k, m^k) = \lambda_{\text{trav}}(\mathbf{x}^k, m^k)) + \lambda_{\text{coll}}(\mathbf{x}^k, m^k),
\end{equation}
or, more generally, a weighted combination to account for the relative importance of terrain and collision hazards.

\subsubsection{Traversability} Traversability captures the likelihood of accidents due to terrain, primarily affecting ground robots. 
We define the expected accident rate per unit distance as
\begin{equation}
\label{eq:trav_cost}
\lambda_{\text{trav}}(\mathbf{x}^k, m^k) := p(\text{accident} \mid \mathbf{x}^k, \gamma(m^k)),
\end{equation}
which can be estimated from experimental studies or manufacturer specifications. We consider a constant value per terrain type and robot.
Nevertheless, note that this formulation can be readily extended to incorporate additional semantic features or advanced risk-assessment metrics for traversability~\cite{dixit2024step, frey23fast}. 

\subsubsection{Collision}  
Collision captures the likelihood of accidents due to proximity to obstacles. This is particularly relevant for aerial platforms, which are sensitive to control errors, wind gusts, or other uncertainties when operating near surfaces. %For a platform $k$ at position $\mathbf{x}^k$, 
We define the collision-related accident rate using a logistic function,
\begin{align}
\lambda_{\text{coll}}(\mathbf{x}^k, m^k) :&= p(\text{accident} \mid \mathbf{x}^k, d(m^k)) \\ &= \frac{1}{1 + e^{\beta (d(m^k) - d_{0.5})}},
\end{align}
where $\beta$ controls the decay of risk with distance, and $d_{0.5}$ is the distance at which the probability of collision reaches 0.5.

\subsection{Heterogeneous Vehicle Routing}
Lastly, we need to compute a solution for the robots to perform all the inspection tasks efficiently.
We formulate this as an asymmetric, capacitated Heterogeneous Vehicle Routing Problem. 
The VRP represents the tasks $v \in V$ as nodes and edges $e \in E$ weighted by the costs, $c^k_{ij},$ of the different robots to move from one to another, as described before.
The goal is to find routes for agents $k \in K$ that depart from a depot, visit all locations exactly once, and return to the depot. 
Asymmetry is introduced by omitting return costs to the depot. 
Each agent has a capacity $C^k$ that limits the cumulative route cost, and to balance workload, we adopt a min–max objective, minimizing the maximum route cost rather than the total cost. 
Altogether yields the following optimization problem,
%To model heterogeneous traversal capabilities, edge costs are defined per agent, denoted $c^k_{ij}$, yielding the following formulation:
\begin{subequations}
\begin{gather}
    \label{eq:HetVRP}
    \min_{x_{ij}^k}\max_{k\in K}\sum_{i} \sum_{j} c_{ij}^{k}x^{k}_{ij}, \quad s.t. \\
    \label{eq:HetVRP1}
    \sum_{k} \sum_{i} x^{k}_{ij}=1 \quad\quad \forall j \in V \setminus \{0\}, \\
    \label{eq:HetVRP2}
    \sum_{k} \sum_{j} x^{k}_{ij}=1 \quad\quad \forall i \in V \setminus \{0\}, \\
    \label{eq:HetVRP3}
    \sum_{k} \sum_{i} x^{k}_{i0}=\sum_{k} \sum_{j} x^{k}_{0j}=\left|{K}\right|, \\
    \label{eq:HetVRP4}
    \sum_{i,j \in S} x^k_{i,j} \leq \left|{S}\right|-1, \quad \forall S \subset V \setminus \{0\},S\neq\emptyset, \\
    \label{eq:HetVRP5}
    \sum_i \sum_j c_{ij}^k x_{ij}^k \leq C^k, \quad \forall k\in K \\
    \label{eq:HetVRP6}
    x^{k}_{ij} \in \{ 0,1 \} \quad\quad \forall i,j \in V,
\end{gather}
\end{subequations}
where $x^k_{ij}$ are binary decision variables indicating whether agent $k$ traverses from $i$ to $j$. Constraints \eqref{eq:HetVRP1}–\eqref{eq:HetVRP2} enforce that each location is visited exactly once, \eqref{eq:HetVRP3} ensures all routes start and end at the depot, \eqref{eq:HetVRP4} eliminates sub-tours that do not include the depot, \eqref{eq:HetVRP5} imposes capacity limits used to prevent infeasible routes, and \eqref{eq:HetVRP6} enforces binary variables. Importantly, limiting each agent's capacity enables constraining routes through infeasible paths due to safety costs.

The resulting paths are sent to each robot, where the platforms' navigation stacks handle low-level motion execution and local collision avoidance during task execution.

\begin{figure}[!t]
    \centering
    \vspace{2mm}
    \includegraphics[width=0.95\linewidth]{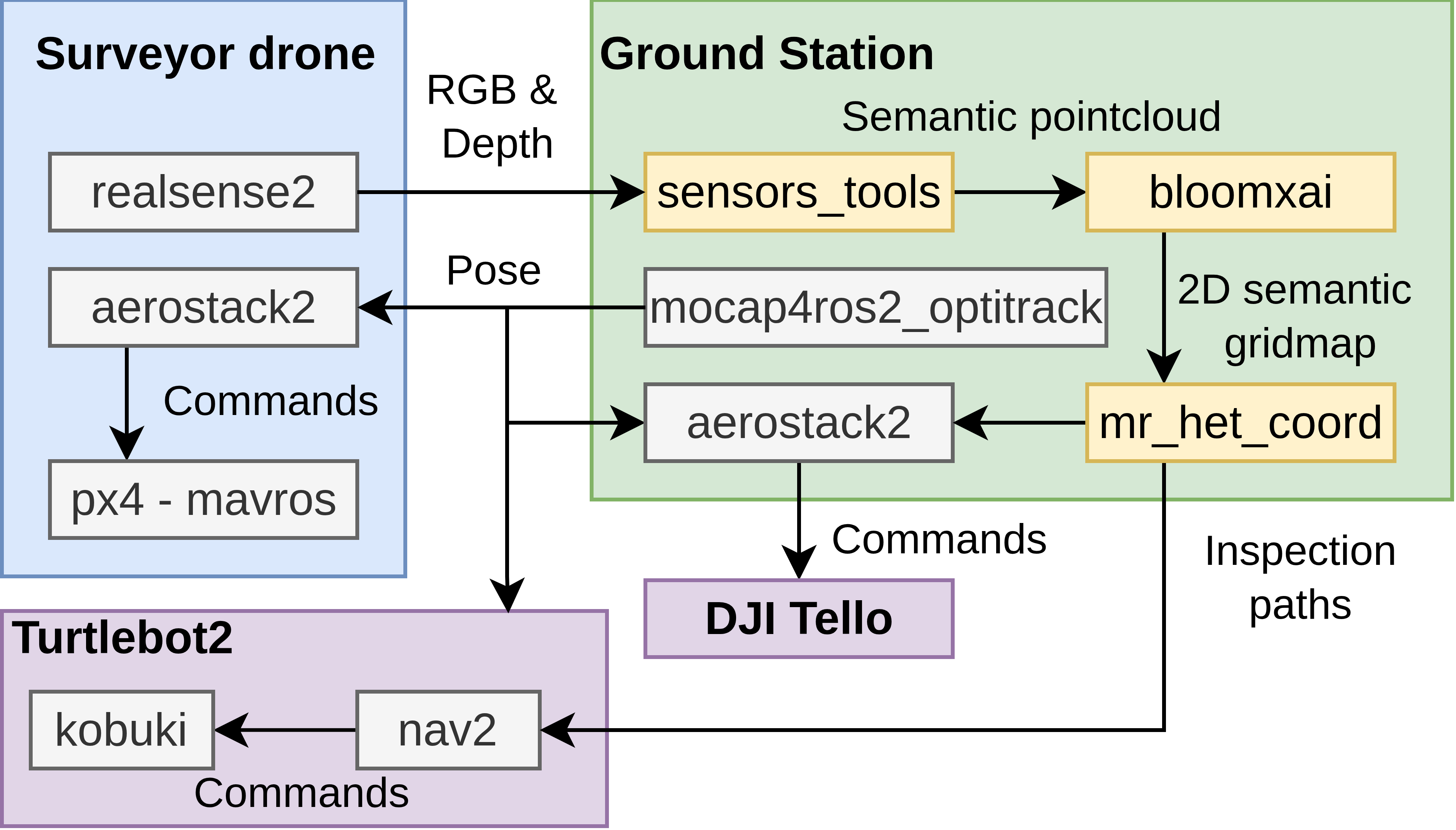}
    \caption{Diagram of the ROS 2 system implementation provided, indicating the components in our framework. Yellow means the component was fully developed for this article. Gray means the component was adapted and integrated into our system.}
    \label{fig:implementation}
\end{figure}

\section{Implementation}
\label{sec:implementation}

We implement our heterogeneous coordination framework in a modular architecture, illustrated in Fig.~\ref{fig:implementation}, where our components are shown in yellow and integrated third-party tools in gray. Although specific packages are used in our implementation, the interfaces are generic and interchangeable (e.g., GPS can replace motion-capture–based localization).

We consider a surveyor drone, localized in space and equipped with an RGB camera and a depth sensor (e.g., RGB-D or learning-based depth estimation~\cite{wang2025vggt}), is used to acquire visual data. Semantic features are extracted through \mbox{\emph{sensors\_tools}}, which integrates Trident~\cite{shi2025trident} for class probability vectors and RayFront’s semantic encoder~\cite{alama2025rayfronts} for raw feature embeddings.

Semantic observations are projected into 3D using depth and sensor pose. Our metric-semantic voxel map, \emph{bloomxai}, extends Bonxai~\cite{faconti2025bonxai}, which uses voxel hashing for efficient large-scale occupancy mapping. Probabilities are fused via Bayesian fusion with regularization~\cite{morilla2023robust}, while raw embeddings are averaged. The resulting 3D representation is processed into a 2D grid map for planning.

Planning is handled by \mbox{\emph{mr\_het\_coord}}, which integrates PRM construction with RRT* and A* over the 2D grid. Routing is formulated as a heterogeneous extension of the classical VRP and solved using the industry-grade OR-Tools solver~\cite{ortools}. The formulation incorporates per-agent cost models while retaining support for common VRP extensions such as time windows and task compatibility constraints.

All components are orchestrated within a ROS~2 framework and deployed in Docker containers to ensure reproducibility and portability across hardware platforms. Platform-specific navigation and execution are delegated to established stacks, Nav2~\cite{macenski2020nav2} for ground robots and Aerostack2~\cite{fernandezcortizas2024aerostack2} for aerial vehicles, providing standardized interfaces for control and navigation. This containerized, modular design enables scalable coordination of heterogeneous robot teams while simplifying integration of new platforms and capabilities.
\section{Experiments \& Results}

We evaluate our heterogeneous coordination framework through extensive virtual experiments and a real-world multi-robot deployment. The virtual experiments demonstrate, under varying scenarios, the benefits of our heterogeneous coordination framework for safer and more efficient planning compared to existing routing solutions~\cite{david2022sweep, zhou2023racer}. The real-world experiment validates the complete pipeline, from semantic mapping to task execution, showing the practicality of the proposed system under real sensing, planning, and execution constraints.

\subsection{Virtual experiment}

\begin{table}[!t]
\centering
\vspace{2mm}
\caption{Parameters of the virtual maps and time required to obtain the distance and costs matrices. We report the average and standard deviation over 5 runs. The planner is only executed at the start, scales well to a large number of goals and enables online replanning in case of map updates since only local parts need to be recomputed.}
\label{tab:virtual_maps_params}
\setlength{\tabcolsep}{3pt}
\begin{tabular}{@{}llllll@{}}
\toprule
Map     & Size [$\SI{}{\meter}$] & \begin{tabular}[c]{@{}l@{}}Num\\ Tasks\end{tabular} & \begin{tabular}[c]{@{}l@{}}Build \\ PRM [$\SI{}{\milli\second}$]\end{tabular} & \begin{tabular}[c]{@{}l@{}}Compute\\ dist. [$\SI{}{\milli\second}$]\end{tabular} & \begin{tabular}[c]{@{}l@{}}Compute het. \\ costs [$\SI{}{\milli\second}$]\end{tabular} \\ \midrule
Orchard &    $24\times12$          & 110                                                 & $2.4\pm0.5$                                                   & $167.2\pm7.3$                                                    & $95.2\pm3.3$                                                           \\ \midrule
Forest  &     $32.5\times17.5$         & 58                                                  & $2.6\pm0.5$                                                   & $31.6\pm1.51$                                                    & $28.6\pm1.1$                                                           \\ \midrule
Park    &     $44.5\times34$          & 94                                                  & $4.4\pm0.8$                                                   & $118.4\pm1.8$                                                    & $36.0\pm1.6$                                                            \\ \midrule
Cave    &      $280\times180$         & 266                                                 & $17.6\pm3.6$                                                  & $3590.2\pm28.7$                                                  & $873.2\pm17.7$                                                         \\ \bottomrule
\end{tabular}
\end{table}

\begin{figure}[!t]
     \centering
     \begin{subfigure}[]{0.22\textwidth}
         \centering
         \includegraphics[width=\textwidth]{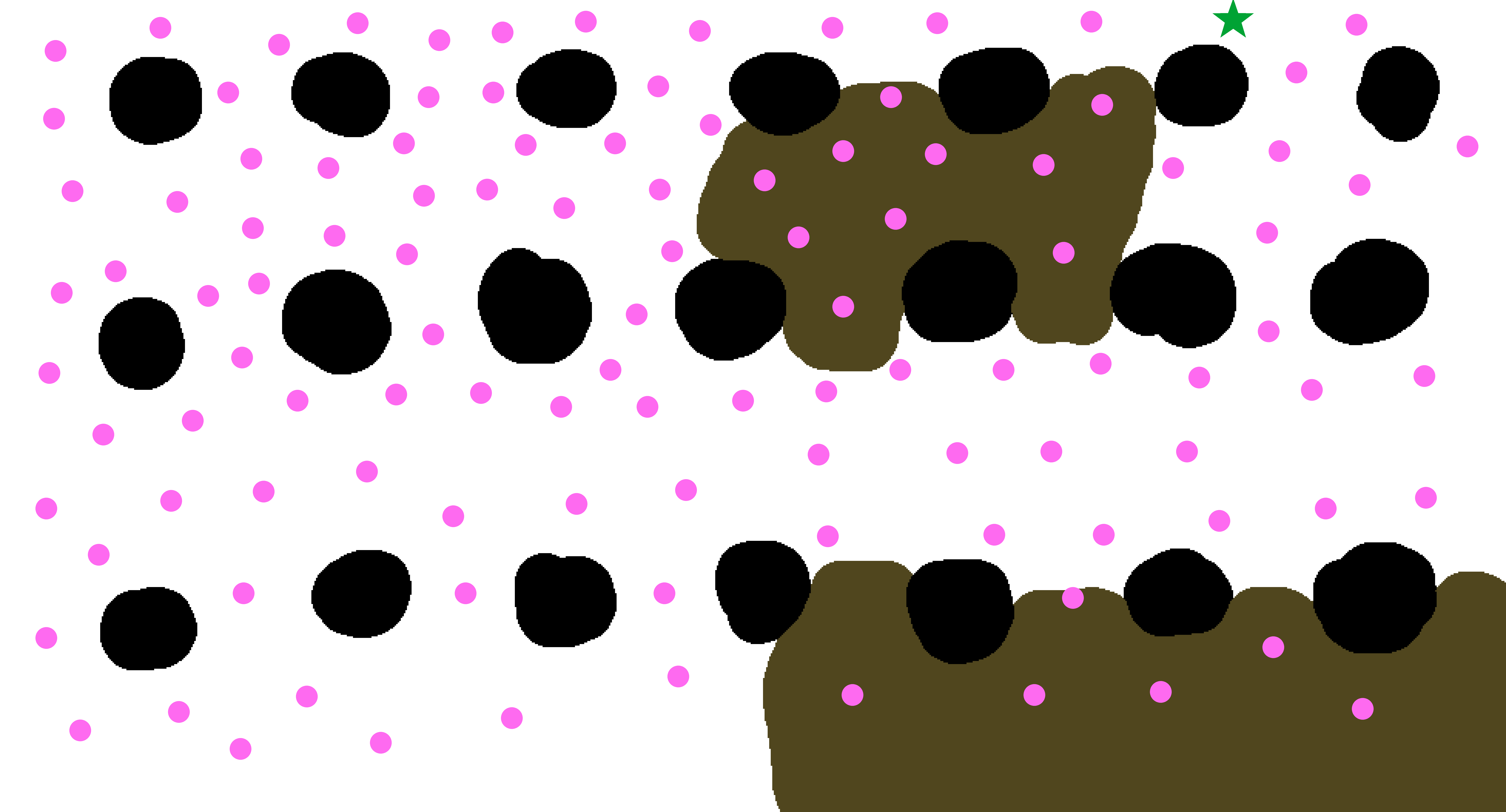}
         \caption{Orchard}
         \label{fig:orchard_virt}
     \end{subfigure}
     \hspace{10pt}
     \begin{subfigure}[]{0.22\textwidth}
         \centering
         \includegraphics[width=\textwidth]{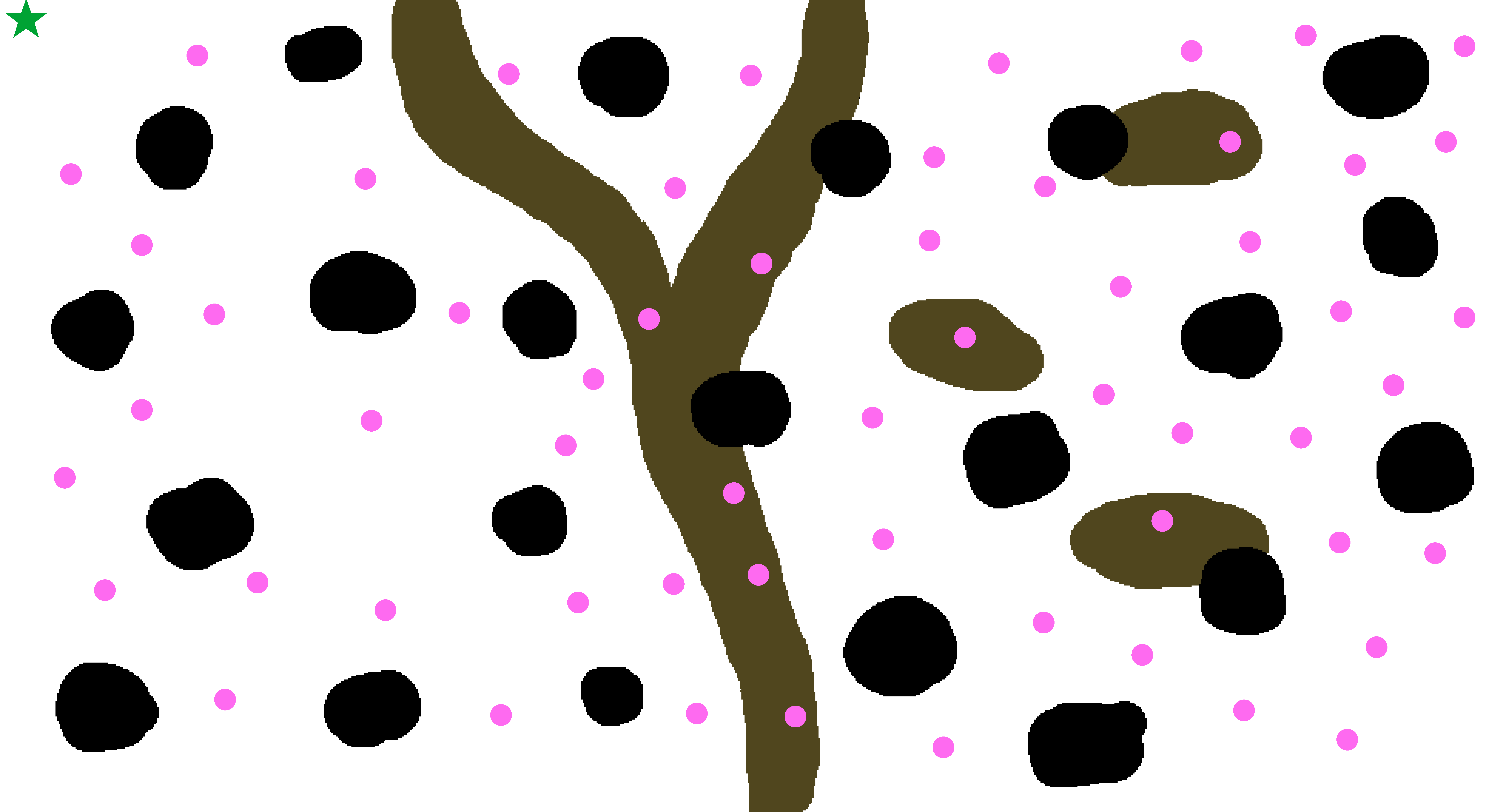}
         \caption{Forest}
         \label{fig:forest_virt}
     \end{subfigure}
     \begin{subfigure}[]{0.1625\textwidth}
         \centering
         \includegraphics[width=\textwidth]{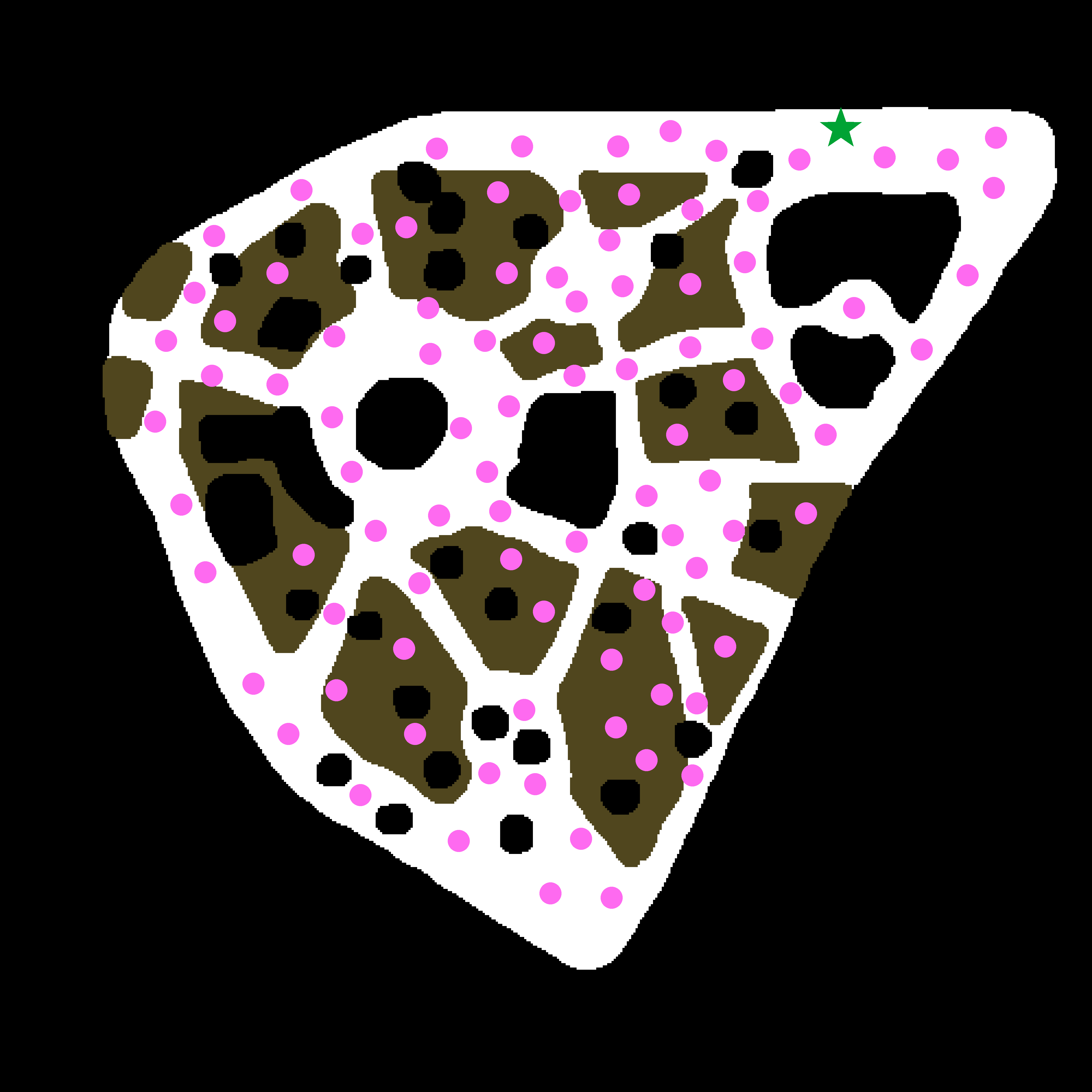}
         \caption{Park}
         \label{fig:park_virt}
     \end{subfigure}
     \hspace{5pt}
          \begin{subfigure}[]{0.25\textwidth}
         \centering
         \includegraphics[width=\textwidth]{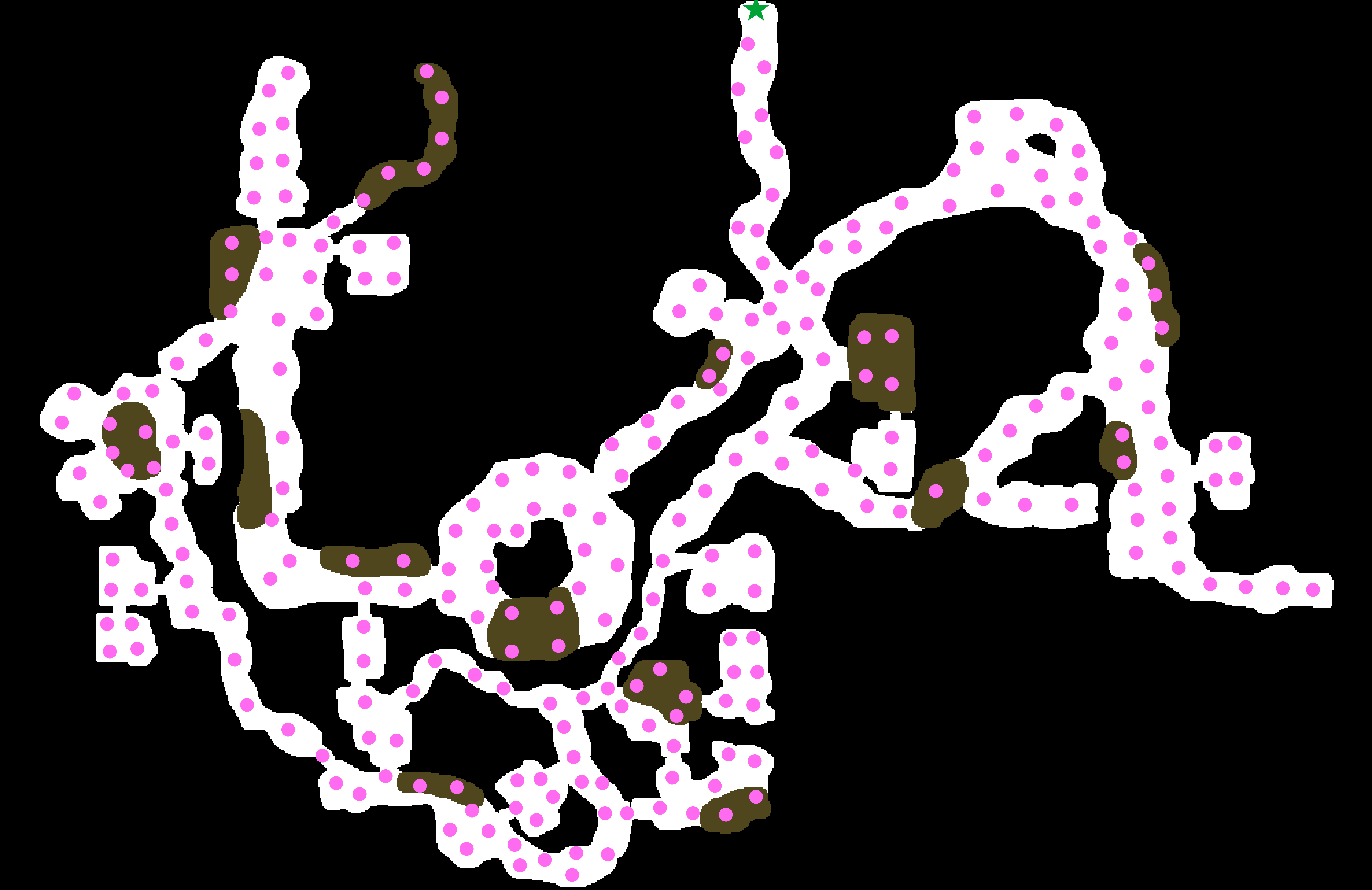}
         \caption{Cave}
         \label{fig:cave_virt}
     \end{subfigure}
        \caption{Virtual environments with variable sizes and complexity. Black represents obstacles, brown represents regions of limited traversability. Pink dots are the tasks, and red stars are the starting positions for the robot team.}
        \label{fig:maps_virt}
        \vspace{-4mm}

\end{figure}

\begin{figure}[!t]
    \centering
    \vspace{2mm}
    \includegraphics[width=0.95\linewidth]{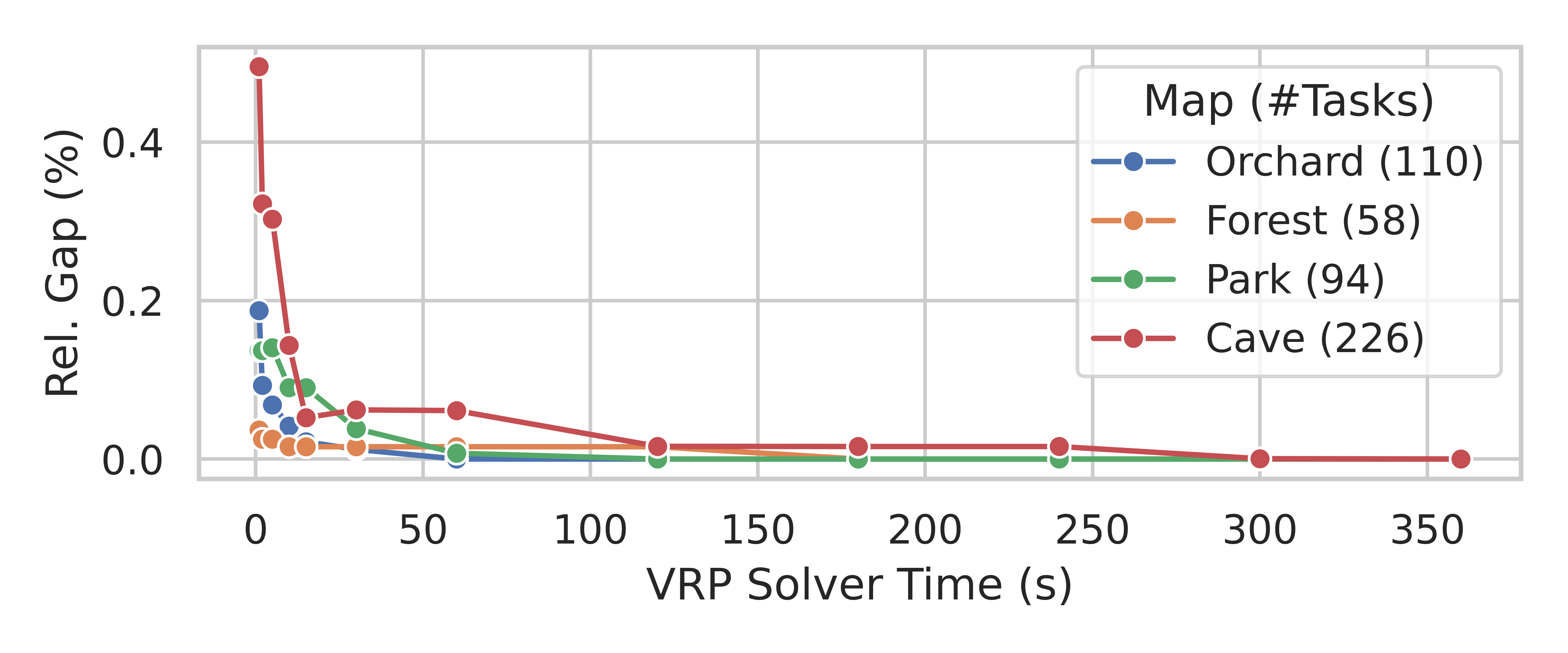}
    \caption{Relative gap for the resulting cost of the VRP solution to the converged solution against the time budget of the OR-Tools solver with respect to $\SI{1}{\second}$ for a team with one drone and one ground robot. Since the solver can be initialized from a previous solution, the maximum time for planning is below $\SI{2}{\minute}$.}
    \label{fig:cost_vs_time}
    \vspace{-5mm}
\end{figure}

The virtual experiments assume a known 2D grid map to validate the proposed heterogeneous cost modeling and planning components across environments of varying size and complexity (Fig.~\ref{fig:maps_virt}). In these maps, white cells denote free space, black cells represent obstacles, and gray regions indicate terrain with limited traversability for ground robots. The selected environments reflect representative inspection scenarios, including areas that are inaccessible to certain platforms due to terrain or narrow passages. The dimensions of each map and the number of inspection tasks are summarized in Table~\ref{tab:virtual_maps_params}.

We first measure the time required to construct the PRM graph and compute distances and costs for the full VRP. As shown in Table~\ref{tab:virtual_maps_params}, this step completes well under a minute, even for large and complex maps.
The OR-Tools solver is combinatorial and heuristic, with a global initialization phase followed by a time-limited local search. We evaluate solver performance for varying local search durations on all virtual maps using a drone–ground robot team. As shown in \reffig{fig:cost_vs_time}, solutions reach over 90\% of the maximum performance within a few seconds for most maps. Since the solver can initialize from a previous solution, only the first planning iteration requires the full computation, typically under \SI{2}{\minute}. These results indicate that our planning framework is efficient even for a large number of tasks.
For the remaining experiments, we set the solver time to \SI{120}{\second} to ensure consistent convergence across maps.

We then solve the inspection mission paths for the different maps using different robot configurations for aerial (A) and ground (G) platforms of 2 and 4. We compare our method against the standard distance-based formulation used in the literature~\cite{david2022sweep, zhou2023racer}. For each map and team configuration, we report the maximum planned distance and time, as those define the total mission performance; and the worst case accident probability for traversability and collision.

\begin{table}[!t]
\centering
\vspace{2mm}
\caption{Results for the virtual experiments. Worst-case (maximum) team performance is reported. Homogeneous (Hom) is the standard method used in the literature for coordination~\cite{david2022sweep, zhou2023racer}. Heterogeneous (Het) corresponds to our proposed planning.}
\label{tab:virtual_results}
\setlength{\tabcolsep}{6pt}
\begin{tabular}{@{}llcccc@{}}
\toprule
Metric & Method / Team & Orchard & Forest & Park & Cave \\ \midrule

\multirow{4}{*}{\begin{tabular}[c]{@{}l@{}}Distance [m] $\downarrow$\end{tabular}}
& Hom (G,A)   & \textbf{80}  & \textbf{86}  & \textbf{148} & \textbf{1542} \\
& \textbf{Het (G,A)}   & 115 & 116 & 233 & 1989 \\
\cmidrule{2-6}
& Hom (2G,2A) & \textbf{45}  & \textbf{43}  & \textbf{75}  & \textbf{696} \\
& \textbf{Het (2G,2A)} & 66  & 63  & 120 & 1195 \\ \midrule \midrule

\multirow{4}{*}{\begin{tabular}[c]{@{}l@{}}Time [s] $\downarrow$\end{tabular}}
& Hom (G,A)   & 160 & 172 & 295 & 3083 \\
& \textbf{Het (G,A)}   & \textbf{115} & \textbf{163} & \textbf{237} & \textbf{2098} \\
\cmidrule{2-6}
& Hom (2G,2A) & 90  & 86  & 149 & 1391 \\
& \textbf{Het (2G,2A)} & \textbf{68}  & \textbf{75}  & \textbf{126} & \textbf{1195} \\ \midrule \midrule

\multirow{4}{*}{\begin{tabular}[c]{@{}l@{}}Traversability \\ accident \\ probability  $\downarrow$\end{tabular}}
& Hom (G,A)   & 0.16 & 0.61 & 0.91 & 1.00 \\
& \textbf{Het (G,A)}   & \textbf{0.07} & \textbf{0.00} & \textbf{0.11} & \textbf{0.04} \\
\cmidrule{2-6}
& Hom (2G,2A) & 0.24 & 0.40 & 0.68 & 0.97 \\
& \textbf{Het (2G,2A)} & \textbf{0.10} & \textbf{0.00} & \textbf{0.11} & \textbf{0.01} \\ \midrule \midrule

\multirow{4}{*}{\begin{tabular}[c]{@{}l@{}}Collision \\ accident \\ probability  $\downarrow$\end{tabular}}
& Hom (G,A)   & 0.37 & 0.67 & 0.45 & 0.70 \\
& \textbf{Het (G,A)}   & \textbf{0.05} & \textbf{0.29} & \textbf{0.12} & \textbf{0.12} \\
\cmidrule{2-6}
& Hom (2G,2A) & 0.24 & 0.63 & 0.22 & 0.64 \\
& \textbf{Het (2G,2A)} & \textbf{0.02} & \textbf{0.12} & \textbf{0.06} & \textbf{0.09} \\
\bottomrule
\vspace{-8mm}
\end{tabular}
\end{table}

We set the ground robot’s terrain-dependent accident rate $\lambda_{\text{trav}}$ in Eq.~\refeq{eq:trav_cost} 
to $0.05$ for difficult terrain and \num{1e-5} for all other terrain classes\footnote{A ratio of $0.05$ represents 1 accidents per \SI{20}{\meter}, and \num{1e-5} one per \SI{100}{\kilo\meter}}. Obstacle-related safety costs are set with $\beta=10$ and $d_{0.5}=\SI{0.5}{\meter}$. This configuration penalizes ground robots traversing rough terrain while favoring task assignments near obstacles when safe, and restricts drones from flying close to obstacles unless necessary. 

Table~\ref{tab:virtual_results} reports the resulting distances, traversal times, and accident probabilities for each platform and map. Compared to the standard distance-based formulation~\cite{david2022sweep, zhou2023racer}, our heterogeneous planning method achieves higher efficiency in traversal time, despite increasing the total distance. This improvement is due to incorporating platform-specific velocities when computing paths. Moreover, the resulting routes exhibit lower accident probabilities, both from terrain-related traversability and proximity to obstacles, demonstrating that our cost-aware formulation effectively balances mission speed and safety across heterogeneous robots. Noticeably, the accident probability can be configured in how much we value risk of the mission. Increasing the importance of the safety costs in risk averse situations can reduce the accident probability to $0$ when possible, returning that the mission is not feasible if no routes are found. The following real experiment shows this behavior.

\begin{figure*}[!t]
    \centering
    \vspace{2mm}
    \includegraphics[width=0.85\linewidth]{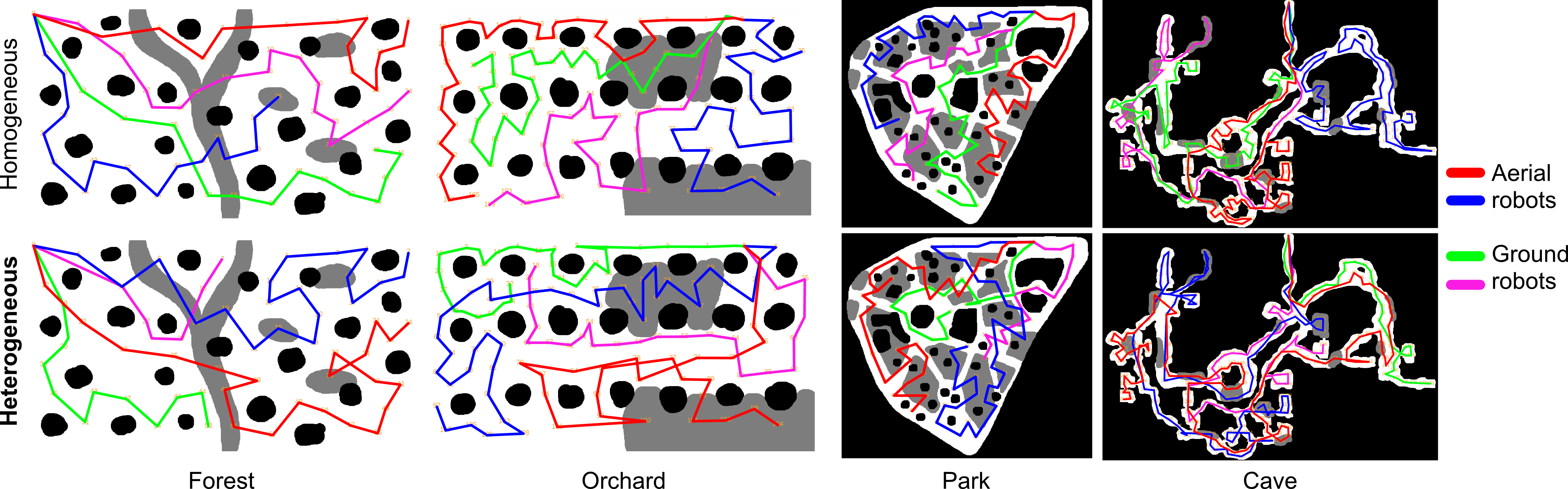}
    \caption{Results for planning with standard homogeneous planning (top) and \textbf{our heterogeneous} method (bottom) for virtual environments. Thanks to considering platform-specific costs, ground robots minimize traversing through rough terrain, and aerial robots attend inspection tasks far from obstacles.}
    \label{fig:virtual_result}
    \vspace{-4mm}
\end{figure*}

\subsection{Real-world experiment}

We validated our framework in a real-world inspection mission with a heterogeneous team of robots, shown in \reffig{fig:teaser}, using the system architecture in \reffig{fig:implementation}. The inspection area spans $6 \times 5.5~\si{\meter}$ and contains five inspection targets represented by animal plushes. Tall objects were placed as obstacles, while smaller elements such as cables and pebbles were used to impede ground robot traversability without constituting hard obstacles.

A surveyor drone equipped with a downward-facing RealSense D435i RGB-D camera performed a reconnaissance flight following a lawn-mower trajectory. The drone uses a Pixhawk flight controller with an Nvidia Orin Nano companion computer and is localized using an OptiTrack motion capture system. RGB-D streams were processed in real time on a ground station with an Nvidia RTX4080 GPU to extract open-vocabulary semantic features using Trident. The text queries used for semantic extraction are shown in Figure~\ref{fig:diagram} and were expanded using OpenAI ImageNet templates. 
While semantic inference and map construction are offloaded due to the surveyor drone’s onboard compute constraints, recent work has shown that comparable semantic mapping pipelines can run fully onboard robots equipped with more powerful hardware~\cite{alama2025rayfronts}.

\begin{figure}[!t]
    \centering
    \includegraphics[width=0.8\linewidth]{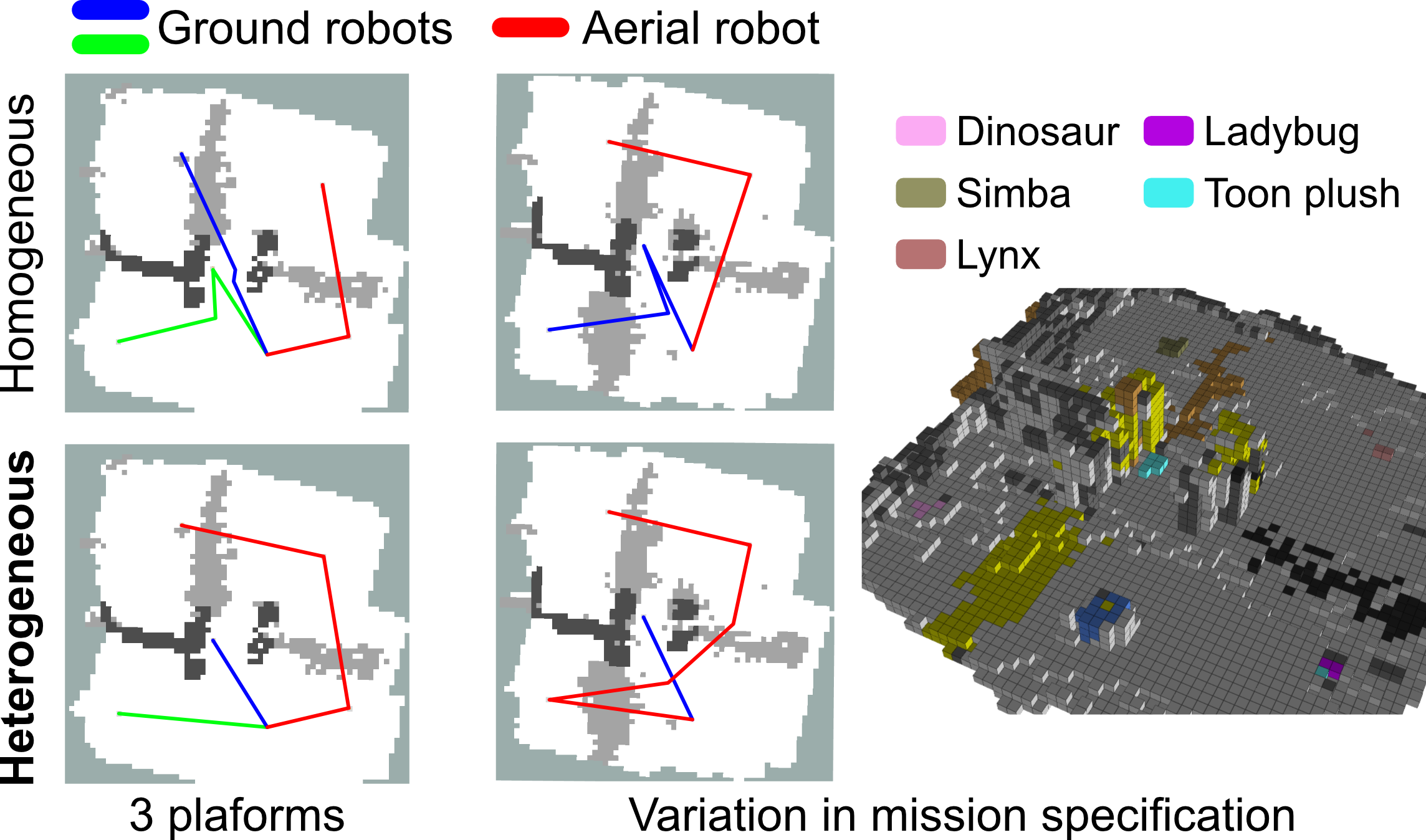}
    \caption{Results for variations in mission specification in the real experiment. (Left) Using two ground platforms instead of one results in our method still avoiding traversal through rough terrain, with the aerial platform assuming more tasks due to its higher velocity. (Right) Fine-grained task and terrain specifications enable rich and flexible mission definition.}
    \label{fig:variation}
    \vspace{-4mm}
\end{figure}

For this deployment, traversability costs $\lambda_{\text{trav}}$ for the ground platform were set to $5$ for the classes \emph{pebbles} and \emph{cables}, effectively preventing traversal of these regions due to capacity limits in the heterogeneous routing formulation. Inspection tasks were defined from the \emph{animal} class. The resulting traversability and signed distance field maps used for cost evaluation are shown in Figure~\ref{fig:diagram}. The Heterogeneous Vehicle Routing Problem was solved on the ground station with a $3~\si{\second}$ planning horizon to generate paths for each robot.

The inspection team consisted of a Turtlebot2 with an Intel NUC and a DJI Tello controlled remotely from the ground station. Nominal planning velocities were set to $0.25$ and $0.5~\si{\meter\per\second}$ for the ground and aerial platforms, respectively. Both robots relied on OptiTrack for localization and executed the planned routes through their respective navigation stacks. As shown in Figure~\ref{fig:diagram}, the resulting solution balances workload between the platforms, assigning $20~\si{\second}$ to the ground robot and $14~\si{\second}$ to the drone, while avoiding traversability-impaired regions for the ground platform and narrow corridors for the aerial robot. Actual execution times, after a short path processing delay from the platforms, were around $25~\si{\second}$ for the ground robot and $10~\si{\second}$ for the drone. A video of the deployment is provided in the supplementary material.
This deployment demonstrates the framework’s capability to safely and efficiently coordinate heterogeneous robots in real environments, leveraging RGB-D perception for informed, platform-aware planning.

Finally, we evaluated variations in mission specification derived from the same metric-semantic map, with results shown in \reffig{fig:variation}. On the left, we demonstrate a mission involving two ground robots and one drone, resulting in a shorter, balanced mission while still respecting platform-specific constraints. On the right, we illustrate fine-grained task definitions using individual semantic classes (\emph{dinosaur, simba, lynx, toon plush}), excluding \emph{ladybug}, and treating \emph{cardboard} as traversability-impeding terrain. These variations highlight the flexibility of our semantic scene understanding in enabling rich, adaptable mission definitions without modifying the underlying planning framework.
\section{Conclusion and Future Work}
\label{sec:conclusion}

In this work, we present a semantic-aware coordination framework for heterogeneous multi-robot inspection that integrates environment understanding with platform-aware routing. Leveraging metric-semantic maps built from RGB-D perception and open-vocabulary vision models, our system identifies arbitrary inspection tasks, estimates platform-specific traversability and collision costs, and computes safe, optimized routes through a Heterogeneous Vehicle Routing Problem formulation. Extensive simulations demonstrated the efficiency and safety improvements of our approach compared to conventional distance-based homogeneous coordination, while the real-world deployment validated its practical feasibility in coordinating aerial and ground platforms. Our framework extends the possibilities for coordinated multi-robot applications with a modular, open-source solution integrating semantic perception with heterogeneous multi-robot planning. Future work will focus on extending the framework to outdoor and larger-scale environments, supporting more complex mission specifications, and exploring learning-based approaches to automatically infer inspection tasks and traversal costs from high-level platform and mission descriptions.

\balance

\bibliography{tex/references}
\bibliographystyle{IEEEtran} % use IEEEtran.bst style
\end{document}